\documentclass[10pt,twocolumn,letterpaper]{article}

\usepackage[pagenumbers]{cvpr} 

\usepackage[dvipsnames]{xcolor}

\definecolor{cvprblue}{rgb}{0.21,0.49,0.74}
\usepackage[pagebackref,breaklinks,colorlinks,citecolor=cvprblue]{hyperref}
\usepackage[accsupp]{axessibility} 
\def\paperID{9116} 
\def\confName{CVPR}
\def\confYear{2024}
\newcommand{\mytabular}[2]{\centering\scalebox{#1}{#2}}

\title{Scaling Up Video Summarization Pretraining with Large Language Models}

\author{Dawit Mureja Argaw$^{1,2}$ \quad 
Seunghyun Yoon$^2$ \quad
Fabian Caba Heilbron$^2$ \quad
Hanieh Deilamsalehy$^2$ \quad \\
Trung Bui$^2$ \quad
Zhaowen Wang$^2$ \quad
Franck Dernoncourt$^2$ \quad
Joon Son Chung$^1$ \\ \\
$^1$  KAIST \quad\quad  $^2$ Adobe Research}

\begin{document}
%
%
%
%
%
%
%
\newcommand{\ba}{{\mathbf{a}}}
\newcommand{\bb}{{\mathbf{b}}}
\newcommand{\bc}{{\mathbf{c}}}
\newcommand{\bd}{{\mathbf{d}}}
\newcommand{\bolde}{{\mathbf{e}}}
\newcommand{\boldf}{{\mathbf{f}}}
\newcommand{\bg}{{\mathbf{g}}}
\newcommand{\bh}{{\mathbf{h}}}
\newcommand{\bi}{{\mathbf{i}}}
\newcommand{\bj}{{\mathbf{j}}}
\newcommand{\bk}{{\mathbf{k}}}
\newcommand{\bl}{{\mathbf{l}}}
\newcommand{\bm}{{\mathbf{m}}}
\newcommand{\bn}{{\mathbf{n}}}
\newcommand{\bo}{{\mathbf{o}}}
\newcommand{\bp}{{\mathbf{p}}}
\newcommand{\bq}{{\mathbf{q}}}
\newcommand{\br}{{\mathbf{r}}}
\newcommand{\bs}{{\mathbf{s}}}
\newcommand{\bt}{{\mathbf{t}}}
\newcommand{\bu}{{\mathbf{u}}}
\newcommand{\bv}{{\mathbf{v}}}
\newcommand{\bw}{{\mathbf{w}}}
\newcommand{\bx}{{\mathbf{x}}}
\newcommand{\by}{{\mathbf{y}}}
\newcommand{\bz}{{\mathbf{z}}}
\newcommand{\model}{{MODEL}}

\newcommand{\bA}{\mathbf{A}}
\newcommand{\bB}{\mathbf{B}}
\newcommand{\bC}{\mathbf{C}}
\newcommand{\bD}{\mathbf{D}}
\newcommand{\bE}{\mathbf{E}}
\newcommand{\bF}{\mathbf{F}}
\newcommand{\bG}{\mathbf{G}}
\newcommand{\bH}{\mathbf{H}}
\newcommand{\bI}{\mathbf{I}}
\newcommand{\bJ}{\mathbf{J}}
\newcommand{\bK}{\mathbf{K}}
\newcommand{\bL}{\mathbf{L}}
\newcommand{\bM}{\mathbf{M}}
\newcommand{\bN}{\mathbf{N}}
\newcommand{\bO}{\mathbf{O}}
\newcommand{\bP}{\mathbf{P}}
\newcommand{\bQ}{\mathbf{Q}}
\newcommand{\bR}{\mathbf{R}}
\newcommand{\bS}{\mathbf{S}}
\newcommand{\bT}{\mathbf{T}}
\newcommand{\bU}{\mathbf{U}}
\newcommand{\bV}{\mathbf{V}}
\newcommand{\bW}{\mathbf{W}}
\newcommand{\bX}{\mathbf{X}}
\newcommand{\bY}{\mathbf{Y}}
\newcommand{\bZ}{\mathbf{Z}}

\newcommand{\calA}{{\mathcal{A}}}
\newcommand{\calB}{{\mathcal{B}}}
\newcommand{\calC}{{\mathcal{C}}}
\newcommand{\calD}{{\mathcal{D}}}
\newcommand{\calE}{{\mathcal{E}}}
\newcommand{\calF}{{\mathcal{F}}}
\newcommand{\calG}{{\mathcal{G}}}
\newcommand{\calH}{{\mathcal{H}}}
\newcommand{\calI}{{\mathcal{I}}}
\newcommand{\calJ}{{\mathcal{J}}}
\newcommand{\calK}{{\mathcal{K}}}
\newcommand{\calL}{{\mathcal{L}}}
\newcommand{\calM}{{\mathcal{M}}}
\newcommand{\calN}{{\mathcal{N}}}
\newcommand{\calO}{{\mathcal{O}}}
\newcommand{\calP}{{\mathcal{P}}}
\newcommand{\calQ}{{\mathcal{Q}}}
\newcommand{\calR}{{\mathcal{R}}}
\newcommand{\calS}{{\mathcal{S}}}
\newcommand{\calT}{{\mathcal{T}}}
\newcommand{\calU}{{\mathcal{U}}}
\newcommand{\calV}{{\mathcal{V}}}
\newcommand{\calW}{{\mathcal{W}}}
\newcommand{\calX}{{\mathcal{X}}}
\newcommand{\calY}{{\mathcal{Y}}}
\newcommand{\calZ}{{\mathcal{Z}}}
\newcommand{\calbX}{\mbox{\boldmath $\mathcal{X}$}}
\newcommand{\calbY}{\mbox{\boldmath $\mathcal{Y}$}}

\newcommand{\bcalA}{\mbox{\boldmath $\calA$}}
\newcommand{\bcalB}{\mbox{\boldmath $\calB$}}
\newcommand{\bcalC}{\mbox{\boldmath $\calC$}}
\newcommand{\bcalD}{\mbox{\boldmath $\calD$}}
\newcommand{\bcalE}{\mbox{\boldmath $\calE$}}
\newcommand{\bcalF}{\mbox{\boldmath $\calF$}}
\newcommand{\bcalG}{\mbox{\boldmath $\calG$}}
\newcommand{\bcalH}{\mbox{\boldmath $\calH$}}
\newcommand{\bcalI}{\mbox{\boldmath $\calI$}}
\newcommand{\bcalJ}{\mbox{\boldmath $\calJ$}}
\newcommand{\bcalK}{\mbox{\boldmath $\calK$}}
\newcommand{\bcalL}{\mbox{\boldmath $\calL$}}
\newcommand{\bcalM}{\mbox{\boldmath $\calM$}}
\newcommand{\bcalN}{\mbox{\boldmath $\calN$}}
\newcommand{\bcalO}{\mbox{\boldmath $\calO$}}
\newcommand{\bcalP}{\mbox{\boldmath $\calP$}}
\newcommand{\bcalQ}{\mbox{\boldmath $\calQ$}}
\newcommand{\bcalR}{\mbox{\boldmath $\calR$}}
\newcommand{\bcalS}{\mbox{\boldmath $\calS$}}
\newcommand{\bcalT}{\mbox{\boldmath $\calT$}}
\newcommand{\bcalU}{\mbox{\boldmath $\calU$}}
\newcommand{\bcalV}{\mbox{\boldmath $\calV$}}
\newcommand{\bcalW}{\mbox{\boldmath $\calW$}}
\newcommand{\bcalX}{\mbox{\boldmath $\calX$}}
\newcommand{\bcalY}{\mbox{\boldmath $\calY$}}
\newcommand{\bcalZ}{\mbox{\boldmath $\calZ$}}

\newcommand{\sfA}{\mbox{$\mathsf A$}}
\newcommand{\sfB}{\mbox{$\mathsf B$}}
\newcommand{\sfC}{\mbox{$\mathsf C$}}
\newcommand{\sfD}{\mbox{$\mathsf D$}}
\newcommand{\sfE}{\mbox{$\mathsf E$}}
\newcommand{\sfF}{\mbox{$\mathsf F$}}
\newcommand{\sfG}{\mbox{$\mathsf G$}}
\newcommand{\sfH}{\mbox{$\mathsf H$}}
\newcommand{\sfI}{\mbox{$\mathsf I$}}
\newcommand{\sfJ}{\mbox{$\mathsf J$}}
\newcommand{\sfK}{\mbox{$\mathsf K$}}
\newcommand{\sfL}{\mbox{$\mathsf L$}}
\newcommand{\sfM}{\mbox{$\mathsf M$}}
\newcommand{\sfN}{\mbox{$\mathsf N$}}
\newcommand{\sfO}{\mbox{$\mathsf O$}}
\newcommand{\sfP}{\mbox{$\mathsf P$}}
\newcommand{\sfQ}{\mbox{$\mathsf Q$}}
\newcommand{\sfR}{\mbox{$\mathsf R$}}
\newcommand{\sfS}{\mbox{$\mathsf S$}}
\newcommand{\sfT}{\mbox{$\mathsf T$}}
\newcommand{\sfU}{\mbox{$\mathsf U$}}
\newcommand{\sfV}{\mbox{$\mathsf V$}}
\newcommand{\sfW}{\mbox{$\mathsf W$}}
\newcommand{\sfX}{\mbox{$\mathsf X$}}
\newcommand{\sfY}{\mbox{$\mathsf Y$}}
\newcommand{\sfZ}{\mbox{$\mathsf Z$}}

\newcommand{\balpha}{\mbox{\boldmath $\alpha$}}
\newcommand{\bbeta}{\mbox{\boldmath $\beta$}}
\newcommand{\bgamma}{\mbox{\boldmath $\gamma$}}
\newcommand{\bdelta}{\mbox{\boldmath $\delta$}}
\newcommand{\bepsilon}{\mbox{\boldmath $\epsilon$}}
\newcommand{\bvarepsilon}{\mbox{\boldmath $\varepsilon$}}
\newcommand{\bzeta}{\mbox{\boldmath $\zeta$}}
\newcommand{\boldeta}{\mbox{\boldmath $\eta$}}
\newcommand{\btheta}{\mbox{\boldmath $\theta$}}
\newcommand{\bvartheta}{\mbox{\boldmath $\vartheta$}}
\newcommand{\biota}{\mbox{\boldmath $\iota$}}
\newcommand{\bkappa}{\mbox{\boldmath $\kappa$}}
\newcommand{\blambda}{\mbox{\boldmath $\lambda$}}
\newcommand{\bmu}{\mbox{\boldmath $\mu$}}
\newcommand{\bnu}{\mbox{\boldmath $\nu$}}
\newcommand{\bxi}{\mbox{\boldmath $\xi$}}
\newcommand{\bpi}{\mbox{\boldmath $\pi$}}
\newcommand{\bvarpi}{\mbox{\boldmath $\varpi$}}
\newcommand{\brho}{\mbox{\boldmath $\rho$}}
\newcommand{\bvarrho}{\mbox{\boldmath $\varrho$}}
\newcommand{\bsigma}{\mbox{\boldmath $\sigma$}}
\newcommand{\bvarsigma}{\mbox{\boldmath $\varsigma$}}
\newcommand{\btau}{\mbox{\boldmath $\tau$}}
\newcommand{\bupsilon}{\mbox{\boldmath $\upsilon$}}
\newcommand{\bphi}{\mbox{\boldmath $\phi$}}
\newcommand{\bvarphi}{\mbox{\boldmath $\varphi$}}
\newcommand{\bchi}{\mbox{\boldmath $\chi$}}
\newcommand{\bpsi}{\mbox{\boldmath $\psi$}}
\newcommand{\bomega}{\mbox{\boldmath $\omega$}}

\newcommand{\bGamma}{\mbox{\boldmath $\Gamma$}}
\newcommand{\bDelta}{\mbox{\boldmath $\Delta$}}
\newcommand{\bTheta}{\mbox{\boldmath $\Theta$}}
\newcommand{\bLambda}{\mbox{\boldmath $\Lambda$}}
\newcommand{\bXi}{\mbox{\boldmath $\Xi$}}
\newcommand{\bPi}{\mbox{\boldmath $\Pi$}}
\newcommand{\bSigma}{\mbox{\boldmath $\Sigma$}}
\newcommand{\bUpsilon}{\mbox{\boldmath $\Upsilon$}}
\newcommand{\bPhi}{\mbox{\boldmath $\Phi$}}
\newcommand{\bPsi}{\mbox{\boldmath $\Psi$}}
\newcommand{\bOmega}{\mbox{\boldmath $\Omega$}}

\newcommand{\veca}{{\vec{\ba}}}
\newcommand{\vecb}{{\vec{\bb}}}
\newcommand{\vecc}{{\vec{\bc}}}
\newcommand{\vecd}{{\vec{\bd}}}
\newcommand{\vece}{{\vec{\bolde}}}
\newcommand{\vecf}{{\vec{\boldf}}}
\newcommand{\vecg}{{\vec{\bg}}}
\newcommand{\vech}{{\vec{\bh}}}
\newcommand{\veci}{{\vec{\bi}}}
\newcommand{\vecj}{{\vec{\bj}}}
\newcommand{\veck}{{\vec{\bk}}}
\newcommand{\vecl}{{\vec{\bl}}}
\newcommand{\vecm}{{\vec{\bm}}}
\newcommand{\vecn}{{\vec{\bn}}}
\newcommand{\veco}{{\vec{\bo}}}
\newcommand{\vecp}{{\vec{\bp}}}
\newcommand{\vecq}{{\vec{\bq}}}
\newcommand{\vecr}{{\vec{\br}}}
\newcommand{\vecs}{{\vec{\bs}}}
\newcommand{\vect}{{\vec{\bt}}}
\newcommand{\vecu}{{\vec{\bu}}}
\newcommand{\vecv}{{\vec{\bv}}}
\newcommand{\vecw}{{\vec{\bw}}}
\newcommand{\vecx}{{\vec{\bx}}}
\newcommand{\vecy}{{\vec{\by}}}
\newcommand{\vecz}{{\vec{\bz}}}

\newcommand{\vecxi}{{\vec{\bxi}}}
\newcommand{\vecphi}{{\vec{\bphi}}}
\newcommand{\vecvarphi}{{\vec{\bvarphi}}}
\newcommand{\vecbeta}{{\vec{\bbeta}}}
\newcommand{\vecdelta}{{\vec{\bdelta}}}
\newcommand{\vectheta}{{\vec{\btheta}}}

\newcommand{\Real}{\mathbb R}
\newcommand{\Complex}{\mathbb C}
\newcommand{\Natural}{\mathbb N}
\newcommand{\Integer}{\mathbb Z}


\newcommand{\bone}{\mbox{\boldmath $1$}}
\newcommand{\bzero}{\mbox{\boldmath $0$}}
\newcommand{\0}{{\bf 0}}

\newcommand{\be}{\begin{eqnarray}}
\newcommand{\ee}{\end{eqnarray}}
\newcommand{\bee}{\begin{eqnarray*}}
\newcommand{\eee}{\end{eqnarray*}}

\newcommand{\matrixb}{\left[ \begin{array}}
\newcommand{\matrixe}{\end{array} \right]}

\newcommand{\argmax}{\operatornamewithlimits{\arg \max}}
\newcommand{\argmin}{\operatornamewithlimits{\arg \min}}

\newcommand{\mean}[1]{\left \langle #1 \right \rangle}
\newcommand{\ave}{\mathbb E}
\newcommand{\E}{\mathbb E}
\newcommand{\empha}[1]{{\color{red} \bf #1}}
\newcommand{\fracpartial}[2]{\frac{\partial #1}{\partial  #2}}
\newcommand{\incomplete}[1]{\textcolor{red}{#1}}

\def\doublespace{\renewcommand{\baselinestretch}{2}\large\normalsize}
\def\singlespace{\renewcommand{\baselinestretch}{1}\large\normalsize}
\def\onehalfspace{\renewcommand{\baselinestretch}{1.5}\large\normalsize}
\def\onequaterspace{\renewcommand{\baselinestretch}{1.3}\large\normalsize}
\def\threequaterspace{\renewcommand{\baselinestretch}{1.7}\large\normalsize}
\def\smallspace{\renewcommand{\baselinestretch}{-.9}\large\normalsize}
\def\tinyspace{\renewcommand{\baselinestretch}{-.7}\large\normalsize}

\newcommand{\tr} { \textrm{tr} }
\newcommand{\re} { \textrm{re} }
\newcommand{\im} { \textrm{im} }
\newcommand{\diag} { \textrm{diag} }
\newcommand{\ddiag} { \textrm{ddiag} }
\newcommand{\off} { \textrm{off} }
\newcommand{\vectxt} { \textrm{vec} }

\newcommand{\lla}{\left\langle}
\newcommand{\rra}{\right\rangle}
\newcommand{\llbr}{\left\lbrack}
\newcommand{\rrbr}{\right\rbrack}
\newcommand{\llb}{\left\lbrace}
\newcommand{\rrb}{\right\rbrace}


\newcommand{\RR}{I\!\!R} 
\newcommand{\Nat}{I\!\!N} 
\newcommand{\CC}{I\!\!\!\!C} 

\newcommand{\Tref}[1]{Table~\ref{#1}}
\newcommand{\Eref}[1]{Eq.~(\ref{#1})}
\newcommand{\Fref}[1]{Fig.~\ref{#1}}
\newcommand{\FCref}[1]{Chapter.~\ref{#1}}
\newcommand{\Sref}[1]{Sec.~\ref{#1}}
\newcommand{\Aref}[1]{Algo.~\ref{#1}}

\def\eg{\emph{e.g.}}
\def\Eg{\emph{E.g.}}
\def\etal{\emph{et al.}}
\def\ie{\emph{i.e.}}

\maketitle
\begin{abstract}
Long-form video content constitutes a significant portion of internet traffic, making automated video summarization an essential research problem. However, existing video summarization datasets are notably limited in their size, constraining the effectiveness of state-of-the-art methods for generalization. Our work aims to overcome this limitation by capitalizing on the abundance of long-form videos with dense speech-to-video alignment and the remarkable capabilities of recent large language models (LLMs) in summarizing long text. We introduce an automated and scalable pipeline for generating a large-scale video summarization dataset using LLMs as Oracle summarizers. By leveraging the generated dataset, we analyze the limitations of existing approaches and propose a new video summarization model that effectively addresses them. To facilitate further research in the field, our work also presents a new benchmark dataset that contains 1200 long videos each with high-quality summaries annotated by professionals. Extensive experiments clearly indicate that our proposed approach sets a new state-of-the-art in video summarization across several benchmarks.
\end{abstract}    
\vspace{-4mm}
\section{Introduction}
\label{sec:intro}
In the current era of information, long-form video content constitutes a significant portion of internet traffic. Consequently, developing models for automated \textit{video summarization} has become an essential research topic~\cite{elhamifar2012see,lu2014bag,zhao2014quasi,lee2012discovering,narasimhan2021clip,zhu2020dsnet,zhao2021reconstructive,jiang2022joint,ji2019video,sharghi2017query,zhao2018hsa}. Video summarization involves automatically creating a condensed summary video from a longer input video, highlighting the key information. This task is highly practical as it allows users to selectively filter the content they wish to explore in greater detail (\eg~promotional trailers) or obtain concise summaries of the content they intend to consume (\eg~recap videos).

Learning to summarize videos, however, is a very ill-posed problem. This is mainly because of the diverse nature of video content and the subjective nature of what constitutes a meaningful summary. 
Therefore, an intuitive, data-driven approach to developing a \textit{robust} video summarizer would involve exposing the model to a large set of video-summary pairs during training.
However, obtaining such a dataset is a daunting and resource-intensive task, primarily due to the manual labor required for annotating summary videos.
This challenge is reflected in existing video summarization datasets like TVSum~\cite{song2015tvsum} and SumMe~\cite{gygli2014creating} which are characterized by a notably small number of video-summary pairs, with only 50 and 25 pairs, respectively. Consequently, state-of-the-art video summarization methods~\cite{narasimhan2021clip,ji2019video,he2023align,narasimhan2022tl} tend to overfit to a specific video domain and their ability to generalize effectively is significantly limited.

The main focus of this work is to address these limitations. Motivated by the abundance of long-form videos with dense \textit{speech-to-video} alignment~\cite{miech2019howto100m} and the recent achievements of large language models (LLMs)~\cite{openai2023gpt,touvron2023llama,manyika2023overview} in comprehending and summarizing extensive textual content, we propose an \textit{automatic} and \textit{scalable} pipeline for large-scale video summarization pretraining. Our key idea is to leverage LLMs as Oracle summarizers to transfer their capabilities from text to the video domain. This enables the scaling up of visual summarization datasets, facilitating the training of video summarizers for scenarios where narration or text is not available.

Given a long narrated video, we first use a speech-to-text model~\cite{radford2023robust,bain2023whisperx} to obtain the textual transcription of the video. Next, we input the text into the LLM in a format where each sentence in the transcript is accompanied by its corresponding timestamp. We then \textit{prompt} the LLM to output an extractive summary of the video transcript by selecting only the most critical and informative moments from the video while maintaining the timestamp and original wording of the selected sentences. The main reason for providing this particular instruction is to guarantee that the extracted textual summary can be seamlessly associated with the corresponding video segments. Finally, we carefully map the extracted textual summary back to the relevant video segments. This process results in a sequence of clips that, when aggregated, form a pseudo-ground truth visual summary. Following this methodology, we create a large-scale dataset, named \textit{Long-form Video Summarization Pretraining} (\textbf{LfVS-P}) dataset, consisting of 250K video-summary pairs for training a robust video summarization model. 

Leveraging the extensive dataset we have generated, our work conducts an analysis of various video summarization baselines. A common approach in most existing works is to frame video summarization as a binary classification~\cite{narasimhan2021clip} where each moment is classified as summary or not, or as frame importance prediction~\cite{narasimhan2022tl,ji2019video,he2023align}, estimating the likelihood of each frame being part of a summary. However, these approaches present two key limitations. Firstly, they suffer from a long-tail distribution problem, characterized by a significant class imbalance, as the number of summary moments in a video is considerably smaller compared to non-summary moments. Secondly, the decision of whether a video segment at a given time step is a summary or not happens independently, without consideration of what was previously classified as a summary, as predictions are made in parallel. This eventually leads to numerous repetitive moments being categorized as summary.

To address these limitations, our work proposes a new video summarization model. We adopt a regression-based approach in which the model decodes continuous feature representations of the summary moments, as opposed to predicting discrete binary classes or importance scores, in order to mitigate the long-tail distribution problem. Furthermore, we utilize an autoregressive decoding process, where the decoding at a given time step $t$ is conditioned on the summary moments decoded up to time $t-1$ and the input video. This sequential scheme enables the network to learn intricate contextual dependencies between summary moments during the generation of a summary. 

We design a Transformer-based~\cite{vaswani2017attention} encoder-decoder architecture that takes a long video as input and autoregressively generates a short summary video. We approach video summarization as a multi-modal problem integrating both visual and textual (transcribed speech) cues, from the input video to guide the prediction of summary videos. Recognizing the prevalence of videos without narration or language cues, we train our framework to depend solely on visual cues when textual information is absent. Consequently, during inference, our model is versatile and can be deployed on videos, whether or not they come with accompanying text.

We conduct comprehensive experiments covering aspects such as problem formulation, network design, and scaling effects. Our results clearly indicate the benefits of large-scale pretraining using the collected dataset for robust cross-dataset generalization. Furthermore, to assess the effectiveness of video summarization models and to encourage ongoing research in this field, we introduce a new benchmark known as \textit{Long-form Video Summarization Testing} (\textbf{LfVS-T}). This benchmark comprises a collection of 1,200 diverse videos, each paired with carefully annotated ground truth summaries produced by professional human annotators. We evaluate our model and existing approaches~\cite{narasimhan2021clip, narasimhan2022tl,jiang2022joint,he2023align} using various metrics. Our autoregressive approach outperforms previous works, establishing a new state-of-the-art across multiple benchmarks.
\vspace{-4mm}
\paragraph{Contributions.} Our work brings three main contributions to video summarization research:
\textbf{(1)} We introduce an automatic and scalable mechanism that leverages publicly available long-form videos and LLMs as oracle summarizers to curate the LfVS-P dataset for large-scale video summarization pretraining. \textbf{(2)} We present a new video summarization model that effectively addresses the limitations of previous works and achieves state-of-the-art performance across several benchmarks. \textbf{(3)} To facilitate further research in the field, we introduce a new benchmark dataset named LfVS-T which contains 1,200 publicly available long videos with high-quality summaries annotated by humans.

\begin{figure*}[!t]
    \centering
    \includegraphics[width=1\linewidth,trim={4.35cm 7.45cm 8.4cm 3.3cm},clip]{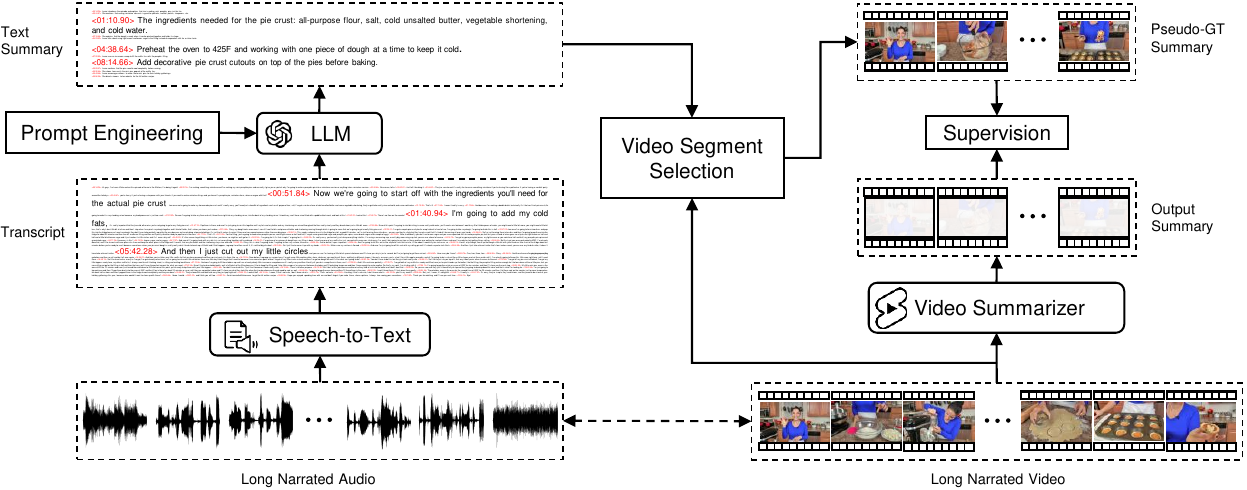}
    \vspace{-6mm}
    \caption{\textbf{Scalable Dataset for Video Summarization}. Given a long-form video with dense speech-to-video alignment, we first use a speech-to-text model~\cite{bain2023whisperx} to transcribe the video. Next, we preprocess the text so that each sentence in the transcript is accompanied by its corresponding start timestamp. We then prompt an LLM~\cite{openai2023gpt,touvron2023llama} to extract the most critical and informative moments from the video along with their timestamp. After extracting the textual summary, we map it back to the relevant video segments to compose a pseudo-ground truth summary. Following this pipeline, we generate a large-scale dataset of video-summary pairs for video summarization pretraining.}
    \label{fig:overview}
    \vspace{-3mm}
\end{figure*}

\vspace{-1mm}
\section{Related Works}
\paragraph{Text Summarization.} 
Text summarization is a fundamental NLP task that aims to generate concise and informative summaries of texts~\cite{alhojely2020recent,gholamrezazadeh2009comprehensive}. Extractive summarization extracts important sentences from the original text~\cite{mihalcea2004textrank,liu2019text}, while abstractive summarization generates new summaries that convey the main points~\cite{lewis2019bart,zhang2020pegasus}. Early works on extractive summarization include TextRank~\cite{mihalcea2004textrank}, an unsupervised graph-based algorithm. Liu~\etal~\cite{liu2019text} enhanced it with contextualized word embeddings~\cite{liu2019text}. The advent of Transformers~\cite{vaswani2017attention} has greatly advanced abstractive summarization, exemplified by models like BART~\cite{lewis2019bart} and Pegasus~\cite{zhang2020pegasus}. Recent progress in text summarization has been driven by the development of large language models (LLMs)~\cite{openai2023gpt,touvron2023llama,manyika2023overview}. These models are able to learn the semantics of the original text and generate summaries that are more informative and comprehensive than traditional summarization models~\cite{zhang2023benchmarking,wang2023element}. Additionally, LLMs have been shown to be capable of generating summaries in a variety of different formats, such as bullet points, paragraphs, and even code~\cite{ouyang2023llm,zhong2023study}. Our work leverages the power of LLMs to create an extensive dataset for video summarization.
\vspace{-3mm}
\paragraph{Video Summarization.} Video summarization is the task of generating a concise representation of a video that captures the main events and ideas. Existing approaches can be broadly categorized as supervised and unsupervised. Many early works focused on unsupervised video summarization~\cite{elhamifar2012see,lu2014bag,zhao2014quasi,lee2012discovering,jung2019discriminative,mahasseni2017unsupervised} partly due to a lack of labeled training datasets. With the emergence of video summarization benchmarks such as SumMe~\cite{gygli2014creating} and TVSum~\cite{song2015tvsum} several supervised approaches~\cite{narasimhan2021clip,zhu2020dsnet,zhao2021reconstructive,jiang2022joint,ji2019video,sharghi2017query,zhao2018hsa,zhang2016video,zhang2018retrospective} have been proposed. Most works focus on generic video summarization~\cite{jiang2022joint,zhao2018hsa,ji2019video} where the most informative moments of an input video are temporally aggregated to compose a summary video. Few other works have explored query-based summarization~\cite{sharghi2017query,narasimhan2021clip,kanehira2018aware,sharghi2016query}, where user-defined natural language queries are used to customize the summaries. Other works have explored a multi-modal setup~\cite{narasimhan2022tl,he2023align,lin2023videoxum,qiu2023multisum} where a text input in the form of video captions or transcribed speech was incorporated along with the video input to guide video summarization. Our work follows a similar formulation and proposes a new video summarization model that attempts to mitigate the limitations of previous approaches.

\section{Scalable Dataset for Video Summarization}
Text summarization has undergone significant advancements in recent years, driven by the exceptional capabilities of large language models (LLMs)~\cite{openai2023gpt,touvron2023llama,manyika2023overview} in comprehending large textual content. In contrast, progress in video summarization has been notably constrained, primarily due to the challenge of obtaining a substantial annotated dataset for the task. Our work aims to bridge this gap by harnessing the power of LLMs to generate a scalable dataset for visual summarization pretraining. The overview of our proposed approach is shown in \Fref{fig:overview}.
\vspace{-2.5mm}
\paragraph{Source Data.} Generating a precise pseudo-ground truth summary from a given video using LLMs as oracle summarizers hinges on the presence of a strong speech-to-visual alignment within the video. For this reason, we make use of the HowTo100M dataset~\cite{miech2019howto100m} which contains more than 1.2M narrated web videos. Given that our work focuses on summarizing long-form videos, we only select videos that are 8 minutes or longer in duration. We then use a state-of-the-art speech-to-text model, Whisper~\cite{bain2023whisperx,radford2023robust}, to transcribe the video. To address potential issues with noisy data curation, we use CLIP embeddings~\cite{radford2021learning} to measure the similarity between the video frames and their corresponding text in the transcript. We subsequently remove videos where the narration lacks sufficient alignment with the visual content in the video.

\begin{figure}[!t]
    \centering
    \includegraphics[width=1\linewidth,trim={8.25cm 11.85cm 14.4cm 3.35cm},clip]{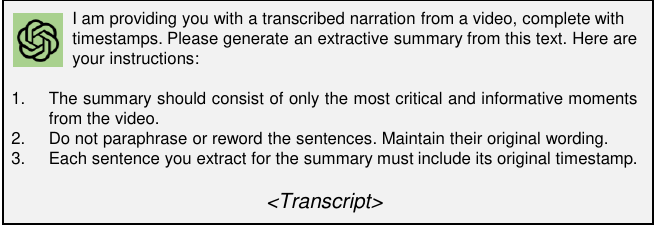}
    \caption{\textbf{Prompt Engineering}. We formulate a prompt instructing an LLM to perform an \emph{extractive text summarization task}. We explicitly emphasize not paraphrasing the wording in the extracted sentences and retaining their timestamps. This ensures seamless matching of the text summary back to the input video.}
    \label{fig:prompt}
    \vspace{-6mm}
\end{figure}


\vspace{-2.5mm}
\paragraph{Prompting LLMs for Extractive Text Summarization.} After obtaining the transcript of a video, we perform text summarization using highly capable LLMs with a large context size~\cite{openai2023gpt,touvron2023llama}. We first preprocess the transcript into a format where each sentence in the text is preceded by its start timestamp, as depicted in \Fref{fig:overview}. This step provides temporal context to the text corpus when feeding it into the LLM, facilitating efficient prompting. We then instruct the LLM to generate an \textit{extractive} summary of the input text by selecting the most crucial and informative moments from the video. Additionally, we instruct the LLM to preserve the original wording of the selected sentences in the summary, along with their corresponding timestamps. This ensures that the extracted textual summary can be seamlessly matched with the respective video segments. The prompt template employed in our work is illustrated in~\Fref{fig:prompt}. We predominantly use GPT-3.5-16K~\cite{openai2023gpt} for large-scale dataset curation. Moreover, we conduct experimental analysis using GPT-4~\cite{openai2023gpt} and Llama 2-13B~\cite{touvron2023llama} as summarizers.

\vspace{-2.5mm}
\paragraph{Pseudo-Ground Truth Video Summary.} We obtain the video segment corresponding to each sentence in the text summary using the \textit{start} and \textit{end} timestamp of the sentence in the transcript. To ensure the accurate selection of video segments that correspond to the text in the summary, thus mitigating any potential timestamp misalignments from the transcription model, we employ a CLIP embedding-based nearest neighborhood search for nearby frames within each summary video segment. Subsequently, the resulting video segments are temporally aggregated to construct a pseudo-ground truth (pGT) summary. Following the pipeline depicted in \Fref{fig:prompt}, we create \textbf{LfVS-P}, a large-scale dataset containing 250K videos and their associated pseudo-ground truth summaries, for video summarization pretraining. In~\Tref{tab:dataset_comparison}, we compare our dataset with existing video summarization datasets. The proposed pretraining dataset stands out for its notable scale and diversity across a wide range of tasks. The longer average video duration in our dataset also facilitates robust training for summarizing videos of varying lengths. 
\vspace{-2mm}

\paragraph{LfVS-T Benchmark.} 
In addition to introducing a large-scale video summarization pretraining dataset, our work establishes a new benchmark, named Long-form Video Summarization Testing (\textbf{LfVS-T}), for evaluating models. 
LfVS-T consists of 1200 videos, each accompanied by manually annotated, high-quality ground truth summaries from professional human annotators. The dataset is sourced from publicly available YouTube content, featuring both narrated and non-narrated videos. The video durations range from 8 to 33 minutes, covering a wide spectrum of 392 distinct categories. 
The size and diversity of LfVS-T (see~\Tref{tab:dataset_comparison}) make it a valuable benchmark for video summarization models, facilitating further research in the field.
\section{Methodology}
\begin{figure*}[!t]
    \centering
    \includegraphics[width=1\linewidth,trim={2.0cm 7.3cm 8.0cm 3.15cm},clip]{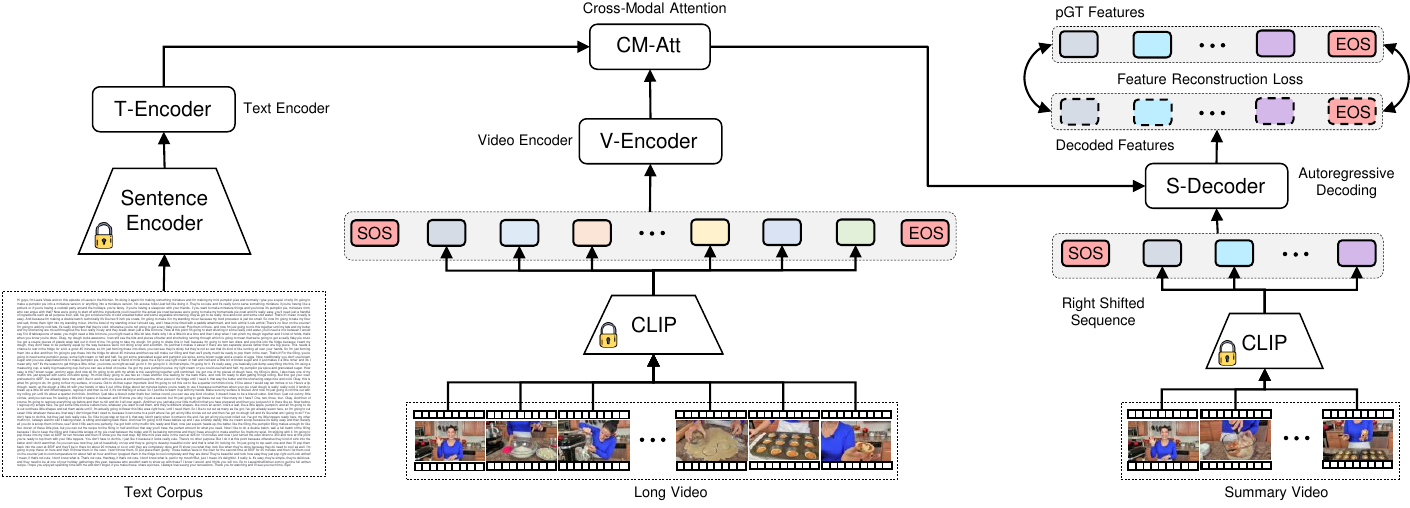}
    \caption{\textbf{Video Summarization Network.} We use a pretrained CLIP~\cite{radford2021learning} model to represent an input video as a sequence of visual tokens. Similarly, we use a pretrained sentence encoder~\cite{liu2019roberta} to encode the long text corpus. In the absence of associated text, we utilize a special \texttt{MASK} token as the text input. We then use a stack of transformer encoders to contextualize the visual and textual features. Next, we incorporate multi-modal cues from the contextualized features via cross-modal attention. Finally, a summary decoder takes the multi-modal features as input and autoregressively decodes the visual representation of the segments that will compose a video summary.}
    \label{fig:model}
    \vspace{-3.5mm}
\end{figure*}

\paragraph{Problem Formulation.}
Video summarization can be purely formulated as a video-to-video problem. However, our experimental observations have indicated a significant performance enhancement when language signal is incorporated.
Therefore, we approach the task as a multi-modal problem, where we take into account both the video content and the text corpus obtained from audio transcription to generate a summary video. Let $V$ denote a video represented as a sequence of frames uniformly sampled every $t$ seconds, \ie~$V=\{X_1, X_2,\ldots, X_n\}$, where $X_n$ denotes a frame at time step $t_n$ and $T$ denote the text data associated with the video represented as a sequence of sentences,~\ie~$T=\{S_1, S_2, \ldots, S_k\}$, where $S_k$ denotes the $k^\mathrm{th}$ sentence. Given $\{V, T\}$ as input, our model outputs a summary video $v = \{Y_1,\ldots,Y_m\}$, where $v \subset V$ and $m \ll n$. We train our network by optimizing the predicted summary $v$ with respect to the pseudo-ground truth summary video $v_\mathrm{pGT}$. Our approach is designed to be flexible to effectively summarize videos, whether they include speech (text) or not, at inference time (refer to \Sref{sec:network}).

\begin{table}[!t]
\begin{center}
    \caption{Comparison with different video summarization datasets.}
    \vspace{-2mm}
    \mytabular{0.73}{
    \begin{tabular}{lcccc}
    \toprule
    Dataset & $\#$ of Videos & $\#$ of Tasks & Avg. Dur. (min) & Annotation \\
    \midrule
    TVSum~\cite{song2015tvsum} & 50 & 10 & 4.2 & Manual\\
    SumMe~\cite{gygli2014creating} & 25 & 25 & 2.4 & Manual\\
    TL:DW?~\cite{narasimhan2022tl} & 12.1K & 185 & 3.1& Automatic\\ 
    \midrule
    LfVS-P (Ours) & 250K & 6.7K & 13.3 & Automatic\\
    LfVS-T (Ours) & 1.2K & 392 & 12.2 & Manual \\
    \bottomrule
    \end{tabular}
    }
    \label{tab:dataset_comparison}
\end{center}
\vspace{-7mm}
\end{table}


\subsection{Video Summarization Network}
\label{sec:network}
We design a Transformer-based~\cite{vaswani2017attention} encoder-decoder network for video summarization. \Fref{fig:model} shows the overview of the proposed model. Our approach consists of four key components: long video encoding, long text encoding, cross-modal attention, and summary video decoding, which are detailed as follows. 
\vspace{-2mm}
\paragraph{Long Video Encoding.} Learning directly from the pixel-space of long videos in an end-to-end manner is often computationally infeasible. Therefore, we opt for using state-of-the-art visual encoders for base feature extraction. Given a long-form video represented as a sequence of frames (sampled every $t$ seconds), we use a pretrained CLIP~\cite{radford2021learning} encoder to obtain a visual embedding for each video frame (see \Eref{eqn:bfe}). This step is equivalent to visual tokenization, wherein we transform an input video into a sequence of feature representations, \ie~$\{x_1, x_2, \ldots, x_n\}$. As shown in \Fref{fig:model}, we augment the visual tokens with special start-of-sequence (\texttt{SOS}) and end-of-sequence (\texttt{EOS}) tokens to mark the beginning and end of the input sequence, respectively. Next, we feed the resulting sequence into a positional encoding layer~\cite{vaswani2017attention} to embed information regarding the relative positions of each token in the sequence. 
\vspace{-1mm}
\begin{equation}
    \{x_1, x_2, \ldots, x_n\} = \mathrm{\textbf{CLIP}}\big(\{X_1, X_2, \ldots, X_n\}\big)
    \label{eqn:bfe}
\end{equation}
After this stage, we pass the resulting sequence to a video encoder (referred to as \textbf{V-Encoder} in \Eref{eqn:context}), which consists of a stack of transformer encoder layers~\cite{vaswani2017attention}. The purpose of the video encoder is to perform temporal reasoning over the input video sequence. In this process, each video moment within the sequence, represented by a visual token, interacts with and attends to every other video moment via a self-attention mechanism. Consequently, the video encoder outputs a sequence of contextualized visual representations, \ie~$\{\hat{x}_0, \hat{x}_1, \hat{x}_2, \ldots, \hat{x}_{n+1}\}$. 
\begin{equation}
    \big\{\hat{x}_i\big\}_{i=0}^{n+1} = \mathrm{\textbf{V-Encoder}}\big(\{\mathrm{\texttt{SOS}}, x_1, \ldots, x_n, \mathrm{\texttt{EOS}}\}\big)
    \label{eqn:context}
    \vspace{-2.5mm}
\end{equation}
\paragraph{Long Text Encoding.} Given a transcribed text associated with an input video, we use a pretrained language model to obtain an encoded representation of the raw text. Considering the text corpus in long videos, where the number of tokens often exceeds the context size of most token-based large language models~\cite{devlin2018bert, liu2019roberta}, we employ a state-of-the-art sentence-based language model, SRoBERTa~\cite{reimers-2019-sentence-bert}, for convenience (\Eref{eqn:bfe_s}). 
\vspace{-1mm}
\begin{equation}
    \{s_1, s_2, \ldots, s_k\} = \mathrm{\textbf{SRoBERTa}}\big(\{S_1, S_2, \ldots, S_k\}\big)
    \label{eqn:bfe_s}
\end{equation}
where $ \{s_1, s_2, \ldots, s_k\}$ denote a sequence of extracted sentence embeddings. To further facilitate text-based contextual learning for video summarization, we pass the extracted sentence embeddings through a text encoder (denoted as \textbf{T-Encoder} in \Eref{eqn:context_s}), which comprises a stack of transformer encoder layers~\cite{vaswani2017attention}. 
\begin{equation}
    \{\hat{s}_1, \hat{s}_2, \ldots, \hat{s}_k\} = \mathrm{\textbf{T-Encoder}}\big(\{s_1,s_2, \ldots, s_k\}\big)
    \label{eqn:context_s}
    \vspace{-1mm}
\end{equation}
 where $\{\hat{s}_1, \hat{s}_2, \ldots, \hat{s}_k\}$ represent the encoded textual representations derived from the text encoder. We aim to design a video summarization framework capable of handling videos with or without corresponding text. To achieve this, we train our model by randomly masking the input text with a special \texttt{MASK} token, using a masking ratio ranging from 0 to 100 percent. This approach allows the network to learn to rely solely on video input when text input is unavailable. At inference, if the input video does not have a corresponding text, we simply use the \texttt{MASK} token as a text input. 
\vspace{-2mm}
\paragraph{Cross-Modal Attention.} To capture inter-modal relationships between video and text inputs, thereby incorporating multi-modal cues for video summarization, we use a cross-modal attention module. Specifically, we adopt the multi-head attention mechanism proposed in \cite{vaswani2017attention} with minor modifications, where we use the encoded visual features as query (\textit{Q}) vector and the encoded text features as key (\textit{K}) and value (\textit{V}) vectors. Let $\hat{x}$ and $\hat{s}$ denote the outputs of the video encoder (\Eref{eqn:context}) and text encoder (\Eref{eqn:context_s}), respectively. The attention mechanism within each head is then defined as follows:
\vspace{-1mm}
\begin{equation}
    \small
    \mathrm{head} = \mathrm{Att.} (\hat{x}W^{Q},\hat{s}W^{K}, \hat{s}W^{V})
\end{equation}
\vspace{-4mm}
\begin{equation}
    \small
    \mathrm{Att.} (Q, K, V) = \mathrm{softmax}(\frac{QK^T}{\sqrt{d_k}})V
\end{equation}
where $W^Q$, $W^K$, and $W^V$ denote learned parameter matrices and $d_k$ is the size of $K$. The cross-modal attention module (denoted as \textbf{CM-Att} in \Eref{eqn:cma}) produces text-conditioned visual features. These features are subsequently utilized as context in a decoder network to generate the video summary.
\vspace{-1mm}
\begin{equation}
    \big\{\hat{x}_{i}^{\hat{s}}\big\}_{i=0}^{n+1} = \mathrm{\textbf{CM-Att}} \big(\big\{\hat{x}_i\big\}_{i=0}^{n+1}, \big\{\hat{s}_j\big\}_{j=1}^{k}\big)
    \label{eqn:cma}
    \vspace{-1mm}
\end{equation}
\paragraph{Summary Video Decoding.}  The summary decoder takes the multi-modal features from the cross-modal attention module as input and generates the visual embeddings of the segments that will compose a video summary in an autoregressive fashion. It comprises a series of transformer decoder layers~\cite{vaswani2017attention}. Similar to next-word prediction in NLP, we implement a next-summary moment prediction scheme in our model. During the training phase, to decode the feature representation of a summary moment at time step $t$ (\ie~$\hat{y}_t$), the summary decoder takes the output of the cross-modal attention module as context and the target pGT summary video sequence up to time step $t-1$ as input as shown in \Eref{eqn:svd}. This design choice accounts for previously selected moments in the summary when choosing the next summary moment from the video. In testing, the summary decoder initiates with the context from the cross-modal attention and the $\mathrm{\texttt{SOS}}$ token. It proceeds to generate the feature representation of the summary video sequence in an autoregressive manner, using the previously generated sequence as input, until the $\mathrm{\texttt{EOS}}$ token is decoded.
\begin{equation}
    \vspace{-1mm}
    \{y_1, y_2, \ldots, y_m\} = \mathrm{\textbf{CLIP}}\big(\{Y_1, Y_2, \ldots, Y_m\}\big)
    \label{eqn:bfe_sum}
\end{equation}
\vspace{-4mm}
\begin{equation}
    \hat{y}_t = \mathrm{\textbf{S-Decoder}} \big(\big\{\hat{x}_{i}^{\hat{s}}\big\}_{i=0}^{n+1}, \{\mathrm{\texttt{SOS}}, y_1, \ldots, y_{t-1}\}\big)
    \label{eqn:svd}
    \vspace{-2mm}
\end{equation}
\paragraph{Training and Inference.} We train our network by optimizing the feature reconstruction loss between the predicted video summary (${\hat{y}_1, \hat{y}_2, \ldots, \hat{y}_m}$) and the pseudo-ground truth summary (${y_1, y_2, \ldots, y_m}$) as follows,
\vspace{-1mm}
\begin{equation}
    \calL = \sum_{i=1}^{m+1} \big|\hat{y}_i - y_i\big|^2
    \label{eqn:loss}
    \vspace{-1mm}
\end{equation}
where $y_{m+1}$ denotes the \texttt{EOS} token. In inference, we utilize nearest neighbor retrieval to match the decoded summary video representations with the CLIP embeddings of the input video sequence. This process selects relevant video moments, which are then temporally aggregated to form the video summary.

\section{Experiment}
\label{sec:experiment}
\paragraph{Implementation Details.} We sample videos at a rate of 1 frame per second (1 fps) to represent input videos and pseudo-ground truth summaries as a sequence of frames. We use CLIP-ViT-L/14~\cite{radford2021learning} for visual tokenization and SRoBERTa-NLI-large~\cite{reimers-2019-sentence-bert} for sentence embedding extraction. Our architecture consists of a video encoder with 6 transformer encoder layers~\cite{vaswani2017attention}, a text encoder with 3 transformer encoder layers, a single cross-modal attention layer, and a summary video decoder with 6 transformer decoder layers~\cite{vaswani2017attention}. Each encoder and decoder layer has a hidden dimension of 1024, 8 attention heads, and a feed-forward dimension of 2048. We train our model using the AdamW optimizer~\cite{loshchilov2017decoupled} with a cosine learning rate annealing strategy~\cite{touvron2021training}, starting from an initial learning rate of $3e-4$. The training utilizes a mini-batch size of 64 and runs for 100 epochs on 4 NVIDIA A6000 GPUs.
\vspace{-2mm}
\paragraph{Evaluation Datasets and Metrics.} We evaluate our approach, along with state-of-the-art video summarization models~\cite{narasimhan2021clip, narasimhan2022tl,jiang2022joint,he2023align}, on established benchmarks such as TVSum~\cite{song2015tvsum} and SumMe~\cite{gygli2014creating}, in addition to the newly introduced LfVS-T benchmark. To measure video summarization performance, we follow established practices~\cite{narasimhan2022tl,jiang2022joint,he2023align} and utilize three different metrics: F1-score, Kendall’s $\tau$~\cite{kendall1945treatment}, and Spearman’s $\rho$~\cite{zwillinger1999crc} metrics.
\begin{table}[!t]
\begin{center}
    \caption{\textbf{Experimental comparison with SoTA approaches}. We train each model on the LfVS-P dataset and evaluate their performance using the proposed LfVS-T benchmark.}
    \vspace{-1mm}
    \mytabular{0.85}{
    \begin{tabular}{lccc}
    \toprule

     Method & F1 Score & $\tau$~\cite{kendall1945treatment} Metric & $\rho$~\cite{zwillinger1999crc} Metric\\
    \midrule
    CLIP-It~\cite{narasimhan2021clip} & 62.87 & 0.129 & 0.225\\
    TL:DW?~\cite{narasimhan2022tl} & \underline{66.25} & 0.138 & 0.233\\
    iPTNet~\cite{jiang2022joint} & 65.80 & 0.140 & 0.237\\
    A2Summ~\cite{he2023align} & 66.04 & \underline{0.143} & \underline{0.246}\\
    Ours & \textbf{68.11} & \textbf{0.158} & \textbf{0.277} \\
    \bottomrule
    \end{tabular}
    }
    \label{tab:results}
\end{center}
\vspace{-6mm}
\end{table}

\subsection{Experimental Results}
\label{sec:results}
We evaluate our approach against several state-of-the-art video summarization models, including CLIP-It~\cite{narasimhan2021clip}, TL:DW?~\cite{narasimhan2022tl}, iPTNet~\cite{jiang2022joint}, A2Summ~\cite{he2023align}. For a fair comparison, we evaluate all methods using the same experimental settings, adhering to their official implementations~\footnote{We reimplement CLIP-It~\cite{narasimhan2021clip} and iPTNet~\cite{jiang2022joint} as their official code is not publicly available.}. To adapt previous models~\cite{narasimhan2021clip, narasimhan2022tl, jiang2022joint, he2023align} formulated for predicting the importance score of each frame (segment) in a video sequence to our experiment setup, we generate ground truth importance scores as follows: we compute the cosine similarity between the CLIP embedding of each video frame $X_i$ (\ie~$x_i$) in the input sequence and the CLIP embedding of every frame in the summary sequence (\ie~$\{y_1, y_2, \ldots, y_m\}$), and assign the maximum value as the importance score $z_i$ for $X_i$ as shown in \Eref{eqn:importance}.
\vspace{-1mm}
\begin{equation}
 \{z_i\}_{i=1}^n = \max_j~s_{i,j},~~~\mathrm{where}~s_{i,j} = \frac{x_i\cdot y_j}{\|x_i\| \|y_j\|},
 \label{eqn:importance}
\end{equation}
A high value of $z_i$ indicates that a video frame $X_i$ is considered important and included in the summary, while a low value suggests that the video moment is dissimilar to any frames in the summary, implying low importance.
\vspace{-2mm}
\paragraph{Comparison with State-of-the-Art.} 
We train our approach and existing video summarization methods on the LfVS-P dataset and evaluate their performances on the proposed LfVS-T benchmark. The results are summarized in~\Tref{tab:results}. As evident from the table, video summarization approaches that integrate text information, such as TL:DW?~\cite{narasimhan2022tl}, A2Summ~\cite{he2023align}, and ours, generally outperform video-only methods like iPTNet~\cite{jiang2022joint}. This is intuitive, as the addition of text information provides extra context, confirming the benefit of framing video summarization as a multi-modal problem. 

As shown in~\Tref{tab:results}, our approach achieves a notably better performance compared to state-of-the-art models across all metrics. For instance, on the F1-score metric, our model outperforms TL:DW? and A2Summ by 2.8\% and 3.1\%, respectively. While previous works predict discrete importance scores for each frame in the input video sequence, our model is designed to decode continuous feature representations of the summary moments. This approach offers benefits in mitigating the inherent class imbalance in video summarization tasks as it allows flexibility in determining how to represent and generate the summary rather than being confined to discrete classes (refer to \Sref{sec:analyses}). More importantly, unlike existing approaches that predict importance scores for all input frames in parallel, our autoregressive model enables conditional generation. Each summary moment is generated based on both the input context and previously generated summary moments, a crucial aspect in video summarization where context is essential for generating subsequent summary moments, contributing to the superior results observed in \Tref{tab:results}. Please refer to the supplementary for further qualitative analysis. 
\vspace{-3mm}
\begin{table}[!t]
\begin{center}
    \caption{\textbf{Results on SumMe and TVSum datasets}. We compare our work and previous methods using the canonical train/test split of SumMe and TVSum datasets. We also conduct cross-dataset generalization experiments by training our model on the LfVS-P dataset and evaluating it on the two datasets.}
    \vspace{-1mm}
    \mytabular{0.73}{
    \begin{tabular}{lcccccc}
    \toprule
     & \multicolumn{3}{c}{SumMe
~\cite{gygli2014creating}} & \multicolumn{3}{c}{TVSum~\cite{song2015tvsum}} \\ \cmidrule(lr){2-4} \cmidrule(lr){5-7} 
     Method & F1 Score & $\tau$~\cite{kendall1945treatment}  & $\rho$~\cite{zwillinger1999crc} & F1 Score & $\tau$~\cite{kendall1945treatment}  & $\rho$~\cite{zwillinger1999crc} \\
    \midrule
    Human~\cite{otani2019rethinking} & 54.00 & 0.205 & 0.213 & 78.00 & 0.177 & 0.204\\ \midrule
    CLIP-It~\cite{narasimhan2021clip}~ & 54.47	& 0.109 &	0.120 & \textbf{66.49} & 0.116 & 0.159 \\
    TL:DW?~\cite{narasimhan2022tl} & 56.46	& 0.111 & 0.128 & 65.84 &0.143 &	0.167 \\
    iPTNet~\cite{jiang2022joint} & 56.61 & 0.114 & 0.131 & \underline{66.16} & 0.148 & 0.174\\
    A2Summ~\cite{he2023align} & \textbf{57.09} & \underline{0.121} & \underline{0.143} & 66.10 & \underline{0.150} & \underline{0.178}\\
    Ours &  \underline{56.94} & \textbf{0.130} & \textbf{0.152} & 66.04 & \textbf{0.155} & \textbf{0.186} \\ \midrule
    \textit{Cross-dataset} & & & & & & \\
    Ours (zero-shot) & 56.72 & 0.125 & 0.148 & 65.76 & 0.151 &	0.182\\
    Ours (fine-tuned) & \textbf{60.42} & \textbf{0.147} & \textbf{0.171} & \textbf{72.38} & \textbf{0.169} &	\textbf{0.203}\\
    \bottomrule
    \end{tabular}
    }
    \label{tab:cross_dataset}
\end{center}
\vspace{-7mm}
\end{table}

\paragraph{Results on SumMe and TVSum Datasets.} In \Tref{tab:cross_dataset}, we evaluate our approach and state-of-the-art methods on SumMe~\cite{gygli2014creating} and TVSum~\cite{song2015tvsum} benchmarks. Following previous works~\cite{narasimhan2021clip,jiang2022joint,he2023align}, we evaluate all methods under canonical train-test splits, conducting experiments five times and reporting the averaged results. As shown in \Tref{tab:cross_dataset}, our approach demonstrates highly competitive, if not superior, performance on both SumMe and TV-Sum datasets. In particular, our approach notably outperforms previous works on Kendall’s $\tau$~\cite{kendall1945treatment}, and Spearman’s $\rho$~\cite{zwillinger1999crc} metrics, which gauge the correlation between predicted and ground truth video summary sequences. These results highlight the benefits of the proposed decoding approach which enables our model to capture sequential dependencies between summary moments when generating a video summary. 

We also conduct cross-dataset generalization experiments, where we train our model on the LfVS-P dataset and evaluate it on SumMe~\cite{gygli2014creating} and TV-Sum~\cite{song2015tvsum} test sets in both zero-shot and fine-tuned settings. As can be seen from \Tref{tab:cross_dataset}, our model, pretrained on pseudo-ground truth summaries, achieves a competitive zero-shot performance on both datasets despite the domain gap. Fine-tuning our pretrained model on the training splits of SumMe and TV-Sum leads to a substantial improvement, establishing a new state-of-the-art in video summarization on both datasets. For instance, on the F1-score metric, our fine-tuned model surpasses the performance of the model trained from scratch by 6.1\% and 9.1\% on the SumMe and TVSum datasets, respectively. This underscores the benefits of our proposed dataset curation framework, which is designed to achieve robust video summarization through large-scale pretraining.
\vspace{-4mm}

\begin{table}[!t]
\begin{center}
    \caption{\textbf{Ablation studies} on LfVS-T benchmark.}
    \vspace{-1.5mm}
    \mytabular{0.7}{
    \begin{tabular}{lccc}
    \toprule

     Method & F1 Score & $\tau$~\cite{kendall1945treatment} Metric & $\rho$~\cite{zwillinger1999crc} Metric\\
    \midrule
    Video (w/o Text Input) & 66.59 & 0.152 & 0.268\\
    w/o Text \& Video Encoder & 62.77 & 0.133 & 0.231\\
    w/o Video Encoder & 63.54 & 0.141 & 0.240 \\
    w/o Text Encoder & 67.49 & 0.154 & 0.272\\
    w/o Cross-Attention & 67.72 & 0.155 & 0.274 \\
    Full Model & \textbf{68.11} & \textbf{0.158} & \textbf{0.277} \\
    \bottomrule
    \end{tabular}
    }
    
    \label{tab:ablation}
\end{center}
\vspace{-8mm}
\end{table}

\subsection{Ablation Studies}
In \Tref{tab:ablation}, we conduct ablation experiments on various network components in our video summarization network. Each model variant is trained on the LfVS-P dataset, and its performance is evaluated on the LfVS-T benchmark.
\vspace{-4mm}
\paragraph{Text Input.} To investigate the significance of incorporating text for video summarization training, we input the contextualized output from the video encoder directly into the summary decoder, omitting the use of a text encoder and cross-modal attention (refer to \Fref{fig:model}). This method frames video summarization as a video-to-video problem, focusing solely on visual information. The results are summarized in \Tref{tab:ablation}. As evident from the table, our video (without text input) baseline performs reasonably well. However, incorporating text input during pretraining to guide video summary generation results in notable improvements. For instance, on the F1 score metric, the baseline trained with text input outperforms the text-less baseline by 2.3\%. The results highlight the benefit of approaching video summarization training as a multi-modal problem, rather than adhering to a pure video-to-video formulation.
\vspace{-4mm}
\paragraph{Video Encoder.} Here, we explore the importance of the video encoder in our framework by directly inputting the sequence visual tokens extracted from a pretrained CLIP~\cite{radford2021learning} model into the cross-modal attention module. As can be inferred from \Tref{tab:ablation}, a baseline without a video encoder significantly underperforms compared to the full model. A similar pattern is observed with a baseline lacking both text and video encoders. This is mainly because the input visual tokens are extracted independently, and feeding them directly to the summary decoder without learning their contextual dependencies via the video encoder provides a much less meaningful context to the summary decoder. Consequently, this leads to a subpar video summarization performance, as shown in \Tref{tab:ablation}.
\vspace{-3mm}
\paragraph{Text Encoder.} To evaluate the effectiveness of text-based contextual learning for video summarization through the text encoder, we train our network by directly inputting the sentence embeddings extracted from a pretrained SRoBERTa~\cite{reimers-2019-sentence-bert} model into the cross-modal attention network. While the baseline performs reasonably well, as shown in \Tref{tab:ablation}, it is evident that a network trained with a text encoder achieves superior performance. This result aligns with our intuition that the text encoder enables the network to learn additional context from the text input for video summarization. 
\vspace{-4mm}
\paragraph{Cross-Modal Attention.} Here, we analyze the benefit of incorporating text and video cues via a cross-modal attention module for decoding video summaries. To achieve this, we concatenate the output of the video and text encoders and input it into the summary decoder. It can be inferred from~\Tref{tab:ablation} that concatenating contextualized text and video features yields competitive performance. This is expected as the decoder attends the different positions in the context sequence when generating each summary moment. However, explicitly performing cross-attention between the video and text features before feeding them to the decoder improves performance. 

\subsection{Experimental Analyses}
\label{sec:analyses}
\paragraph{Problem Formulation.} Our approach frames video summarization as an autoregressive problem, sequentially decoding continuous representations for the summary video. We explore the benefits of this formulation by contrasting it with a classification-based baseline. In the baseline, we substitute the summary decoder in \Fref{fig:model} with a classification layer that categorizes each moment in the input video sequence as a summary or not. This involves feeding the output of the cross-modal attention module into a linear layer and training the network using a cross-entropy loss. The corresponding results are presented in \Tref{tab:analyses}. The classification-based baseline, as seen in the table, significantly underperforms compared to its autoregressive counterpart. Similar to the frame importance score prediction baselines discussed in \Sref{sec:results}, the classification model predicts summary and non-summary moments concurrently, neglecting the sequential dependencies of summary moments. This accounts for its lower performance compared to the autoregressive model in \Tref{tab:analyses}.
\begin{table}[!t]
\begin{center}
    \caption{\textbf{Experimental analyses} on LfVS-T benchmark.}
    \vspace{-2mm}
    \mytabular{0.75}{
    \begin{tabular}{lccc}
    \toprule

     Method & F1 Score & $\tau$~\cite{kendall1945treatment} Metric & $\rho$~\cite{zwillinger1999crc} Metric\\
    \midrule
    \textit{Problem formulation} & & & \\
    Classification & 63.31 & 0.132 & 0.229 \\
    
    Autoregressive & \textbf{68.11} & \textbf{0.158} & \textbf{0.277} \\ \midrule
    
    \textit{Dataset Scale} & & & \\
    10\% & 53.44 & 0.101 & 0.169 \\
    50\% & 64.58 & 0.145 & 0.248 \\
    100\% & \textbf{68.11} & \textbf{0.158} & \textbf{0.277} \\ \midrule
    \textit{LLM (50K samples)} & & & \\
    Llama-2-13B & 44.89 & 0.088 & 0.137 \\
    GPT-3.5-16K & 53.44 & 0.101 & 0.169 \\
    GPT-4 & \textbf{55.96} & \textbf{0.123} & \textbf{0.181} \\ 
    \bottomrule
    \end{tabular}
    }
    \label{tab:analyses}
\end{center}
\vspace{-7.5mm}
\end{table}

\vspace{-3mm}
\paragraph{Dataset Scale.} We investigate the impact of scaling video summarization pretraining by training our network with different amounts of video-pseudo-ground truth summary pairs. Our experiments include using 25K and 125K training samples, accounting for 10\% and 50\%, respectively, of LfVS-P. As evident from \Tref{tab:analyses}, model performance expectedly increases proportionally to the size of the pretraining data. The automatic dataset curation pipeline can introduce noise, affecting the robustness of a model when trained on a small-scale dataset. On the other hand, training on a large-scale dataset provides exposure to diverse samples, contributing to a more robust model, as reflected in the results in \Tref{tab:analyses}.
\vspace{-3mm}
\paragraph{LLM (50K Samples).} The prompt-tuned extractive text summarization using LLMs, illustrated in~\Fref{fig:overview} and~\Fref{fig:prompt}, is a crucial step in our pseudo-ground truth video summary curation process. 
Here, we examine how employing different LLMs as oracle summarizers for generating pretraining data influences video summarization performance. We utilize three state-of-the-art LLMs, Llama 2-13B~\cite{touvron2023llama}, GPT-3.5-16K~\cite{openai2023gpt}, and GPT-4~\cite{openai2023gpt}, to generate 50K training samples (with the same set of input videos for each case) and subsequently train our model. The results are shown in \Tref{tab:analyses}. As can be inferred from the table, a model trained on a dataset obtained using GPT-4 as a summarizer achieves the best performance. This is expected, as GPT-4 has demonstrated superior capabilities in comprehending and summarizing long text corpora~\cite{bubeck2023sparks}, resulting in the generation of high-quality pseudo-ground truth summaries. In contrast, a model trained on a dataset generated using Llama-2-13B~\cite{touvron2023llama} exhibits subpar performance. Our experimental observations indicate that the Llama-2-13B model struggles to precisely follow prompt instructions, leading to the generation of low-quality summaries.

\vspace{-1mm}
\section{Conclusion}
This work introduces an automatic, scalable mechanism using long-form videos and LLMs to create the LfVS-P dataset for large-scale video summarization pretraining. We also propose an autoregressive video summarization model that effectively addresses previous limitations. Additionally, we present the LfVS-T benchmark, comprising 1,200 long videos with human-annotated high-quality summaries. Our extensive comparisons with previous methods demonstrate that our work establishes a new state-of-the-art in video summarization across several benchmarks.

{
    \small
    \bibliographystyle{ieeenat_fullname}
    \bibliography{main}

\begin{thebibliography}{50}
\providecommand{\natexlab}[1]{#1}
\providecommand{\url}[1]{\texttt{#1}}
\expandafter\ifx\csname urlstyle\endcsname\relax
  \providecommand{\doi}[1]{doi: #1}\else
  \providecommand{\doi}{doi: \begingroup \urlstyle{rm}\Url}\fi

\bibitem[Alhojely and Kalita(2020)]{alhojely2020recent}
Suad Alhojely and Jugal Kalita.
\newblock Recent progress on text summarization.
\newblock In \emph{2020 International Conference on Computational Science and Computational Intelligence (CSCI)}, pages 1503--1509. IEEE, 2020.

\bibitem[Bain et~al.(2023)Bain, Huh, Han, and Zisserman]{bain2023whisperx}
Max Bain, Jaesung Huh, Tengda Han, and Andrew Zisserman.
\newblock Whisperx: Time-accurate speech transcription of long-form audio.
\newblock \emph{arXiv preprint arXiv:2303.00747}, 2023.

\bibitem[Bubeck et~al.(2023)Bubeck, Chandrasekaran, Eldan, Gehrke, Horvitz, Kamar, Lee, Lee, Li, Lundberg, et~al.]{bubeck2023sparks}
S{\'e}bastien Bubeck, Varun Chandrasekaran, Ronen Eldan, Johannes Gehrke, Eric Horvitz, Ece Kamar, Peter Lee, Yin~Tat Lee, Yuanzhi Li, Scott Lundberg, et~al.
\newblock Sparks of artificial general intelligence: Early experiments with gpt-4.
\newblock \emph{arXiv preprint arXiv:2303.12712}, 2023.

\bibitem[Devlin et~al.(2018)Devlin, Chang, Lee, and Toutanova]{devlin2018bert}
Jacob Devlin, Ming-Wei Chang, Kenton Lee, and Kristina Toutanova.
\newblock Bert: Pre-training of deep bidirectional transformers for language understanding.
\newblock \emph{arXiv preprint arXiv:1810.04805}, 2018.

\bibitem[Elhamifar et~al.(2012)Elhamifar, Sapiro, and Vidal]{elhamifar2012see}
Ehsan Elhamifar, Guillermo Sapiro, and Rene Vidal.
\newblock See all by looking at a few: Sparse modeling for finding representative objects.
\newblock In \emph{2012 IEEE conference on computer vision and pattern recognition}, pages 1600--1607. IEEE, 2012.

\bibitem[Gholamrezazadeh et~al.(2009)Gholamrezazadeh, Salehi, and Gholamzadeh]{gholamrezazadeh2009comprehensive}
Saeedeh Gholamrezazadeh, Mohsen~Amini Salehi, and Bahareh Gholamzadeh.
\newblock A comprehensive survey on text summarization systems.
\newblock In \emph{2009 2nd International Conference on Computer Science and its Applications}, pages 1--6. IEEE, 2009.

\bibitem[Gygli et~al.(2014)Gygli, Grabner, Riemenschneider, and Van~Gool]{gygli2014creating}
Michael Gygli, Helmut Grabner, Hayko Riemenschneider, and Luc Van~Gool.
\newblock Creating summaries from user videos.
\newblock In \emph{Computer Vision--ECCV 2014: 13th European Conference, Zurich, Switzerland, September 6-12, 2014, Proceedings, Part VII 13}, pages 505--520. Springer, 2014.

\bibitem[He et~al.(2023)He, Wang, Qiu, Bui, Shrivastava, and Wang]{he2023align}
Bo He, Jun Wang, Jielin Qiu, Trung Bui, Abhinav Shrivastava, and Zhaowen Wang.
\newblock Align and attend: Multimodal summarization with dual contrastive losses.
\newblock In \emph{Proceedings of the IEEE/CVF Conference on Computer Vision and Pattern Recognition}, pages 14867--14878, 2023.

\bibitem[Ji et~al.(2019)Ji, Xiong, Pang, and Li]{ji2019video}
Zhong Ji, Kailin Xiong, Yanwei Pang, and Xuelong Li.
\newblock Video summarization with attention-based encoder--decoder networks.
\newblock \emph{IEEE Transactions on Circuits and Systems for Video Technology}, 30\penalty0 (6):\penalty0 1709--1717, 2019.

\bibitem[Jiang and Mu(2022)]{jiang2022joint}
Hao Jiang and Yadong Mu.
\newblock Joint video summarization and moment localization by cross-task sample transfer.
\newblock In \emph{Proceedings of the IEEE/CVF Conference on Computer Vision and Pattern Recognition}, pages 16388--16398, 2022.

\bibitem[Jung et~al.(2019)Jung, Cho, Kim, Woo, and Kweon]{jung2019discriminative}
Yunjae Jung, Donghyeon Cho, Dahun Kim, Sanghyun Woo, and In~So Kweon.
\newblock Discriminative feature learning for unsupervised video summarization.
\newblock In \emph{Proceedings of the AAAI Conference on artificial intelligence}, pages 8537--8544, 2019.

\bibitem[Kanehira et~al.(2018)Kanehira, Van~Gool, Ushiku, and Harada]{kanehira2018aware}
Atsushi Kanehira, Luc Van~Gool, Yoshitaka Ushiku, and Tatsuya Harada.
\newblock Aware video summarization.
\newblock In \emph{Proceedings of the IEEE Conference on Computer Vision and Pattern Recognition}, pages 7435--7444, 2018.

\bibitem[Kendall(1945)]{kendall1945treatment}
Maurice~G Kendall.
\newblock The treatment of ties in ranking problems.
\newblock \emph{Biometrika}, 33\penalty0 (3):\penalty0 239--251, 1945.

\bibitem[Lee et~al.(2012)Lee, Ghosh, and Grauman]{lee2012discovering}
Yong~Jae Lee, Joydeep Ghosh, and Kristen Grauman.
\newblock Discovering important people and objects for egocentric video summarization.
\newblock In \emph{2012 IEEE conference on computer vision and pattern recognition}, pages 1346--1353. IEEE, 2012.

\bibitem[Lewis et~al.(2019)Lewis, Liu, Goyal, Ghazvininejad, Mohamed, Levy, Stoyanov, and Zettlemoyer]{lewis2019bart}
Mike Lewis, Yinhan Liu, Naman Goyal, Marjan Ghazvininejad, Abdelrahman Mohamed, Omer Levy, Ves Stoyanov, and Luke Zettlemoyer.
\newblock Bart: Denoising sequence-to-sequence pre-training for natural language generation, translation, and comprehension.
\newblock \emph{arXiv preprint arXiv:1910.13461}, 2019.

\bibitem[Lin et~al.(2023)Lin, Hua, Chen, Li, Hsiao, Ho, and Luo]{lin2023videoxum}
Jingyang Lin, Hang Hua, Ming Chen, Yikang Li, Jenhao Hsiao, Chiuman Ho, and Jiebo Luo.
\newblock Videoxum: Cross-modal visual and textural summarization of videos.
\newblock \emph{IEEE Transactions on Multimedia}, 2023.

\bibitem[Liu and Lapata(2019)]{liu2019text}
Yang Liu and Mirella Lapata.
\newblock Text summarization with pretrained encoders.
\newblock \emph{arXiv preprint arXiv:1908.08345}, 2019.

\bibitem[Liu et~al.(2019)Liu, Ott, Goyal, Du, Joshi, Chen, Levy, Lewis, Zettlemoyer, and Stoyanov]{liu2019roberta}
Yinhan Liu, Myle Ott, Naman Goyal, Jingfei Du, Mandar Joshi, Danqi Chen, Omer Levy, Mike Lewis, Luke Zettlemoyer, and Veselin Stoyanov.
\newblock Roberta: A robustly optimized bert pretraining approach.
\newblock \emph{arXiv preprint arXiv:1907.11692}, 2019.

\bibitem[Loshchilov and Hutter(2017)]{loshchilov2017decoupled}
Ilya Loshchilov and Frank Hutter.
\newblock Decoupled weight decay regularization.
\newblock \emph{arXiv preprint arXiv:1711.05101}, 2017.

\bibitem[Lu et~al.(2014)Lu, Wang, Mei, Guan, and Feng]{lu2014bag}
Shiyang Lu, Zhiyong Wang, Tao Mei, Genliang Guan, and David~Dagan Feng.
\newblock A bag-of-importance model with locality-constrained coding based feature learning for video summarization.
\newblock \emph{IEEE Transactions on Multimedia}, 16\penalty0 (6):\penalty0 1497--1509, 2014.

\bibitem[Mahasseni et~al.(2017)Mahasseni, Lam, and Todorovic]{mahasseni2017unsupervised}
Behrooz Mahasseni, Michael Lam, and Sinisa Todorovic.
\newblock Unsupervised video summarization with adversarial lstm networks.
\newblock In \emph{Proceedings of the IEEE conference on Computer Vision and Pattern Recognition}, pages 202--211, 2017.

\bibitem[Manyika(2023)]{manyika2023overview}
James Manyika.
\newblock An overview of bard: an early experiment with generative ai, 2023.

\bibitem[Miech et~al.(2019)Miech, Zhukov, Alayrac, Tapaswi, Laptev, and Sivic]{miech2019howto100m}
Antoine Miech, Dimitri Zhukov, Jean-Baptiste Alayrac, Makarand Tapaswi, Ivan Laptev, and Josef Sivic.
\newblock Howto100m: Learning a text-video embedding by watching hundred million narrated video clips.
\newblock In \emph{Proceedings of the IEEE/CVF international conference on computer vision}, pages 2630--2640, 2019.

\bibitem[Mihalcea and Tarau(2004)]{mihalcea2004textrank}
Rada Mihalcea and Paul Tarau.
\newblock Textrank: Bringing order into text.
\newblock In \emph{Proceedings of the 2004 conference on empirical methods in natural language processing}, pages 404--411, 2004.

\bibitem[Narasimhan et~al.(2021)Narasimhan, Rohrbach, and Darrell]{narasimhan2021clip}
Medhini Narasimhan, Anna Rohrbach, and Trevor Darrell.
\newblock Clip-it! language-guided video summarization.
\newblock \emph{Advances in Neural Information Processing Systems}, 34:\penalty0 13988--14000, 2021.

\bibitem[Narasimhan et~al.(2022)Narasimhan, Nagrani, Sun, Rubinstein, Darrell, Rohrbach, and Schmid]{narasimhan2022tl}
Medhini Narasimhan, Arsha Nagrani, Chen Sun, Michael Rubinstein, Trevor Darrell, Anna Rohrbach, and Cordelia Schmid.
\newblock Tl; dw? summarizing instructional videos with task relevance and cross-modal saliency.
\newblock In \emph{European Conference on Computer Vision}, pages 540--557. Springer, 2022.

\bibitem[OpenAI(2023)]{openai2023gpt}
R OpenAI.
\newblock Gpt-4 technical report.
\newblock \emph{arXiv}, pages 2303--08774, 2023.

\bibitem[Otani et~al.(2019)Otani, Nakashima, Rahtu, and Heikkila]{otani2019rethinking}
Mayu Otani, Yuta Nakashima, Esa Rahtu, and Janne Heikkila.
\newblock Rethinking the evaluation of video summaries.
\newblock In \emph{Proceedings of the IEEE/CVF conference on computer vision and pattern recognition}, pages 7596--7604, 2019.

\bibitem[Ouyang et~al.(2023)Ouyang, Zhang, Harman, and Wang]{ouyang2023llm}
Shuyin Ouyang, Jie~M Zhang, Mark Harman, and Meng Wang.
\newblock Llm is like a box of chocolates: the non-determinism of chatgpt in code generation.
\newblock \emph{arXiv preprint arXiv:2308.02828}, 2023.

\bibitem[Qiu et~al.(2023)Qiu, Zhu, Han, Kumar, Mittal, Jin, Yang, Li, Wang, Li, et~al.]{qiu2023multisum}
Jielin Qiu, Jiacheng Zhu, William Han, Aditesh Kumar, Karthik Mittal, Claire Jin, Zhengyuan Yang, Linjie Li, Jianfeng Wang, Bo Li, et~al.
\newblock Multisum: A dataset for multimodal summarization and thumbnail generation of videos.
\newblock \emph{arXiv preprint arXiv:2306.04216}, 2023.

\bibitem[Radford et~al.(2021)Radford, Kim, Hallacy, Ramesh, Goh, Agarwal, Sastry, Askell, Mishkin, Clark, et~al.]{radford2021learning}
Alec Radford, Jong~Wook Kim, Chris Hallacy, Aditya Ramesh, Gabriel Goh, Sandhini Agarwal, Girish Sastry, Amanda Askell, Pamela Mishkin, Jack Clark, et~al.
\newblock Learning transferable visual models from natural language supervision.
\newblock In \emph{International conference on machine learning}, pages 8748--8763. PMLR, 2021.

\bibitem[Radford et~al.(2023)Radford, Kim, Xu, Brockman, McLeavey, and Sutskever]{radford2023robust}
Alec Radford, Jong~Wook Kim, Tao Xu, Greg Brockman, Christine McLeavey, and Ilya Sutskever.
\newblock Robust speech recognition via large-scale weak supervision.
\newblock In \emph{International Conference on Machine Learning}, pages 28492--28518. PMLR, 2023.

\bibitem[Reimers and Gurevych(2019)]{reimers-2019-sentence-bert}
Nils Reimers and Iryna Gurevych.
\newblock Sentence-bert: Sentence embeddings using siamese bert-networks.
\newblock In \emph{Proceedings of the 2019 Conference on Empirical Methods in Natural Language Processing}. Association for Computational Linguistics, 2019.

\bibitem[Sharghi et~al.(2016)Sharghi, Gong, and Shah]{sharghi2016query}
Aidean Sharghi, Boqing Gong, and Mubarak Shah.
\newblock Query-focused extractive video summarization.
\newblock In \emph{Computer Vision--ECCV 2016: 14th European Conference, Amsterdam, The Netherlands, October 11-14, 2016, Proceedings, Part VIII 14}, pages 3--19. Springer, 2016.

\bibitem[Sharghi et~al.(2017)Sharghi, Laurel, and Gong]{sharghi2017query}
Aidean Sharghi, Jacob~S Laurel, and Boqing Gong.
\newblock Query-focused video summarization: Dataset, evaluation, and a memory network based approach.
\newblock In \emph{Proceedings of the IEEE conference on computer vision and pattern recognition}, pages 4788--4797, 2017.

\bibitem[Song et~al.(2015)Song, Vallmitjana, Stent, and Jaimes]{song2015tvsum}
Yale Song, Jordi Vallmitjana, Amanda Stent, and Alejandro Jaimes.
\newblock Tvsum: Summarizing web videos using titles.
\newblock In \emph{Proceedings of the IEEE conference on computer vision and pattern recognition}, pages 5179--5187, 2015.

\bibitem[Touvron et~al.(2021)Touvron, Cord, Douze, Massa, Sablayrolles, and J{\'e}gou]{touvron2021training}
Hugo Touvron, Matthieu Cord, Matthijs Douze, Francisco Massa, Alexandre Sablayrolles, and Herv{\'e} J{\'e}gou.
\newblock Training data-efficient image transformers \& distillation through attention.
\newblock In \emph{International conference on machine learning}, pages 10347--10357. PMLR, 2021.

\bibitem[Touvron et~al.(2023)Touvron, Martin, Stone, Albert, Almahairi, Babaei, Bashlykov, Batra, Bhargava, Bhosale, et~al.]{touvron2023llama}
Hugo Touvron, Louis Martin, Kevin Stone, Peter Albert, Amjad Almahairi, Yasmine Babaei, Nikolay Bashlykov, Soumya Batra, Prajjwal Bhargava, Shruti Bhosale, et~al.
\newblock Llama 2: Open foundation and fine-tuned chat models.
\newblock \emph{arXiv preprint arXiv:2307.09288}, 2023.

\bibitem[Vaswani et~al.(2017)Vaswani, Shazeer, Parmar, Uszkoreit, Jones, Gomez, Kaiser, and Polosukhin]{vaswani2017attention}
Ashish Vaswani, Noam Shazeer, Niki Parmar, Jakob Uszkoreit, Llion Jones, Aidan~N Gomez, {\L}ukasz Kaiser, and Illia Polosukhin.
\newblock Attention is all you need.
\newblock \emph{Advances in neural information processing systems}, 30, 2017.

\bibitem[Wang et~al.(2023)Wang, Zhang, and Wang]{wang2023element}
Yiming Wang, Zhuosheng Zhang, and Rui Wang.
\newblock Element-aware summarization with large language models: Expert-aligned evaluation and chain-of-thought method.
\newblock \emph{arXiv preprint arXiv:2305.13412}, 2023.

\bibitem[Zhang et~al.(2020)Zhang, Zhao, Saleh, and Liu]{zhang2020pegasus}
Jingqing Zhang, Yao Zhao, Mohammad Saleh, and Peter Liu.
\newblock Pegasus: Pre-training with extracted gap-sentences for abstractive summarization.
\newblock In \emph{International Conference on Machine Learning}, pages 11328--11339. PMLR, 2020.

\bibitem[Zhang et~al.(2016)Zhang, Chao, Sha, and Grauman]{zhang2016video}
Ke Zhang, Wei-Lun Chao, Fei Sha, and Kristen Grauman.
\newblock Video summarization with long short-term memory.
\newblock In \emph{Computer Vision--ECCV 2016: 14th European Conference, Amsterdam, The Netherlands, October 11--14, 2016, Proceedings, Part VII 14}, pages 766--782. Springer, 2016.

\bibitem[Zhang et~al.(2018)Zhang, Grauman, and Sha]{zhang2018retrospective}
Ke Zhang, Kristen Grauman, and Fei Sha.
\newblock Retrospective encoders for video summarization.
\newblock In \emph{Proceedings of the European conference on computer vision (ECCV)}, pages 383--399, 2018.

\bibitem[Zhang et~al.(2023)Zhang, Ladhak, Durmus, Liang, McKeown, and Hashimoto]{zhang2023benchmarking}
Tianyi Zhang, Faisal Ladhak, Esin Durmus, Percy Liang, Kathleen McKeown, and Tatsunori~B Hashimoto.
\newblock Benchmarking large language models for news summarization.
\newblock \emph{arXiv preprint arXiv:2301.13848}, 2023.

\bibitem[Zhao and Xing(2014)]{zhao2014quasi}
Bin Zhao and Eric~P Xing.
\newblock Quasi real-time summarization for consumer videos.
\newblock In \emph{Proceedings of the IEEE conference on computer vision and pattern recognition}, pages 2513--2520, 2014.

\bibitem[Zhao et~al.(2018)Zhao, Li, and Lu]{zhao2018hsa}
Bin Zhao, Xuelong Li, and Xiaoqiang Lu.
\newblock Hsa-rnn: Hierarchical structure-adaptive rnn for video summarization.
\newblock In \emph{Proceedings of the IEEE conference on computer vision and pattern recognition}, pages 7405--7414, 2018.

\bibitem[Zhao et~al.(2021)Zhao, Li, Lu, and Li]{zhao2021reconstructive}
Bin Zhao, Haopeng Li, Xiaoqiang Lu, and Xuelong Li.
\newblock Reconstructive sequence-graph network for video summarization.
\newblock \emph{IEEE Transactions on Pattern Analysis and Machine Intelligence}, 44\penalty0 (5):\penalty0 2793--2801, 2021.

\bibitem[Zhong and Wang(2023)]{zhong2023study}
Li Zhong and Zilong Wang.
\newblock A study on robustness and reliability of large language model code generation.
\newblock \emph{arXiv preprint arXiv:2308.10335}, 2023.

\bibitem[Zhu et~al.(2020)Zhu, Lu, Li, and Zhou]{zhu2020dsnet}
Wencheng Zhu, Jiwen Lu, Jiahao Li, and Jie Zhou.
\newblock Dsnet: A flexible detect-to-summarize network for video summarization.
\newblock \emph{IEEE Transactions on Image Processing}, 30:\penalty0 948--962, 2020.

\bibitem[Zwillinger and Kokoska(1999)]{zwillinger1999crc}
Daniel Zwillinger and Stephen Kokoska.
\newblock \emph{CRC standard probability and statistics tables and formulae}.
\newblock Crc Press, 1999.

\end{thebibliography}
}


\end{document}